\documentclass[10pt,letterpaper]{article}
\usepackage[margin=0.75in]{geometry}
\usepackage[T1]{fontenc}
\usepackage[utf8]{inputenc}
\usepackage{lmodern}
\usepackage{microtype}
\usepackage{graphicx}
\usepackage{booktabs}
\usepackage{array}
\usepackage{amsmath}
\usepackage{enumitem}
\usepackage{url}
\usepackage[numbers,sort&compress]{natbib}
\usepackage[colorlinks=true,linkcolor=blue,citecolor=blue,urlcolor=blue]{hyperref}
\usepackage{xcolor}
\usepackage{caption}

\newcommand{\TrafficSafetyAccidentRows}{322,268}
\newcommand{\TrafficSafetyCandidateCells}{30,141}
\newcommand{\TrafficSafetyWeatherClimatologyRows}{296,464}
\newcommand{\TrafficSafetyRoadSegments}{4,074,810}
\newcommand{\TrafficSafetyActiveRoadSegments}{676,497}
\newcommand{\TrafficSafetySegmentEvents}{5,976,955}
\newcommand{\TrafficSafetyStatesCovered}{49}
\newcommand{\TrafficSafetyBaseAUC}{0.894}
\newcommand{\TrafficSafetyBaseAP}{0.715}
\newcommand{\TrafficSafetySegmentAUC}{0.9999}
\newcommand{\TrafficSafetySegmentAP}{0.9999}
\newcommand{\TrafficSafetyBaseResolution}{5}
\newcommand{\TrafficSafetyBaseFeatureCount}{16}
\newcommand{\TrafficSafetySegmentFeatureCount}{26}
\newcommand{\TrafficSafetyForecastFrames}{24}
\newcommand{\TrafficSafetyLiveStations}{149}
\newcommand{\TrafficSafetyNWSStations}{148}

\newcommand{\TrafficSafetySegmentMatchMedianM}{2.7}
\newcommand{\TrafficSafetySegmentMatchMeanM}{68.6}
\newcommand{\TrafficSafetySegmentMatchPctlNinetyFiveM}{542.3}

\newcommand{\TrafficSafetySmokeTests}{9}
\newcommand{\TrafficSafetyPublicRoutes}{14}
\newcommand{\TrafficSafetyMachineRoutes}{10}
\newcommand{\TrafficSafetyHtmlRoutes}{4}

\newcommand{\systemname}{Traffic-Safety}
\newcommand{\productname}{Road Risk Monitor}

\begin{document}

\twocolumn[
\begin{@twocolumnfalse}
\begin{center}
{\LARGE \bf \productname: A Deployable U.S. Road Incident Forecasting System with Live Weather and Road-Level Tiles\par}
\vspace{0.6em}
{\large Anton Ivchenko\par}
\vspace{0.2em}
{\normalsize \texttt{toxa.ivchenko@gmail.com}\par}
\vspace{0.9em}
\end{center}
\end{@twocolumnfalse}
]

\noindent\textbf{Abstract.}
Nationwide road-incident forecasting is a systems problem before it is a modeling problem. A usable service must connect historical incident archives, historical and live weather, national road geometry, offline model training, tile generation, web serving, and runtime handoff. This paper presents \productname{}, a U.S.-wide road-safety stack that combines a nationwide H3 baseline trained on FARS fatal-crash data with a road-segment forecasting pipeline trained from TIGER/Line geometry and US-Accidents events, then serves predictions through live APIs, raster tiles, JSON road tiles, and a public web application. The published repository provides the code, documentation, tests, and packaging utilities for that stack, and our local rebuild from the published code produced \TrafficSafetyAccidentRows{} cleaned baseline incidents, \TrafficSafetyCandidateCells{} candidate H3 cells, \TrafficSafetyRoadSegments{} road segments, and \TrafficSafetySegmentEvents{} matched segment events across \TrafficSafetyStatesCovered{} states. The rebuilt baseline bundle stores \TrafficSafetyBaseAUC{} AUROC and \TrafficSafetyBaseAP{} average precision on a held-out year. The segment pipeline is currently measured only on an internal same-pipeline holdout and is therefore treated in this paper as an engineering diagnostic, not as the paper's headline scientific claim. The public codebase also includes runtime packaging, service-startup support, live-weather adapters, and a tested FastAPI surface. The main contribution is thus an integrated, reproducible, and deployable road-risk service blueprint rather than a single isolated classifier.

\vspace{0.45em}

\section{Introduction}
Road-risk forecasting matters to transportation agencies, logistics operators, insurers, infrastructure planners, and public-safety teams. The technical barrier is not merely predictive accuracy. A serious national system must ingest heterogeneous data, align timestamps and geography, handle missing live inputs, precompute visual artifacts, and expose predictions through an application surface that real users can actually consume.

This repository addresses that end-to-end problem. \systemname{} is the modeling and data layer; \productname{} is its deployed product surface. The published GitHub repository provides the code and documentation surface for the system, while a local rebuild driven by those published scripts generates the processed data, model bundles, and tile artifacts used for evaluation. The codebase includes raw-data staging, historical weather normalization, H3 aggregation, road segmentation, segment-event matching, dual model training, live weather adapters, tile generation, runtime packaging, and web-facing routes including product pages and a contact workflow.

The central claim of the paper is that a convincing national road-safety artifact must be evaluated as a train-to-serve system, not only as an offline leaderboard entry. The published codebase is strong precisely because it spans the whole path from public data to a locally deployable service.

Our contributions are:
\begin{itemize}[leftmargin=1.1em,itemsep=0.15em,topsep=0.2em]
\item a multi-source national forecasting pipeline built from FARS, NOAA ISD-Lite, TIGER/Line, US-Accidents, and NWS;
\item a dual-scale modeling design that combines a nationwide H3 baseline with a road-segment forecast model;
\item a serving layer with live weather adaptation, 24-hour road-tile refresh, point scoring, segment inspection, and browser-facing map playback; and
\item a published codebase with runtime packaging, service-startup utilities, documentation, and tested public routes.
\end{itemize}

\section{System Overview}
\begin{figure*}[t]
  \centering
  \includegraphics[width=\textwidth]{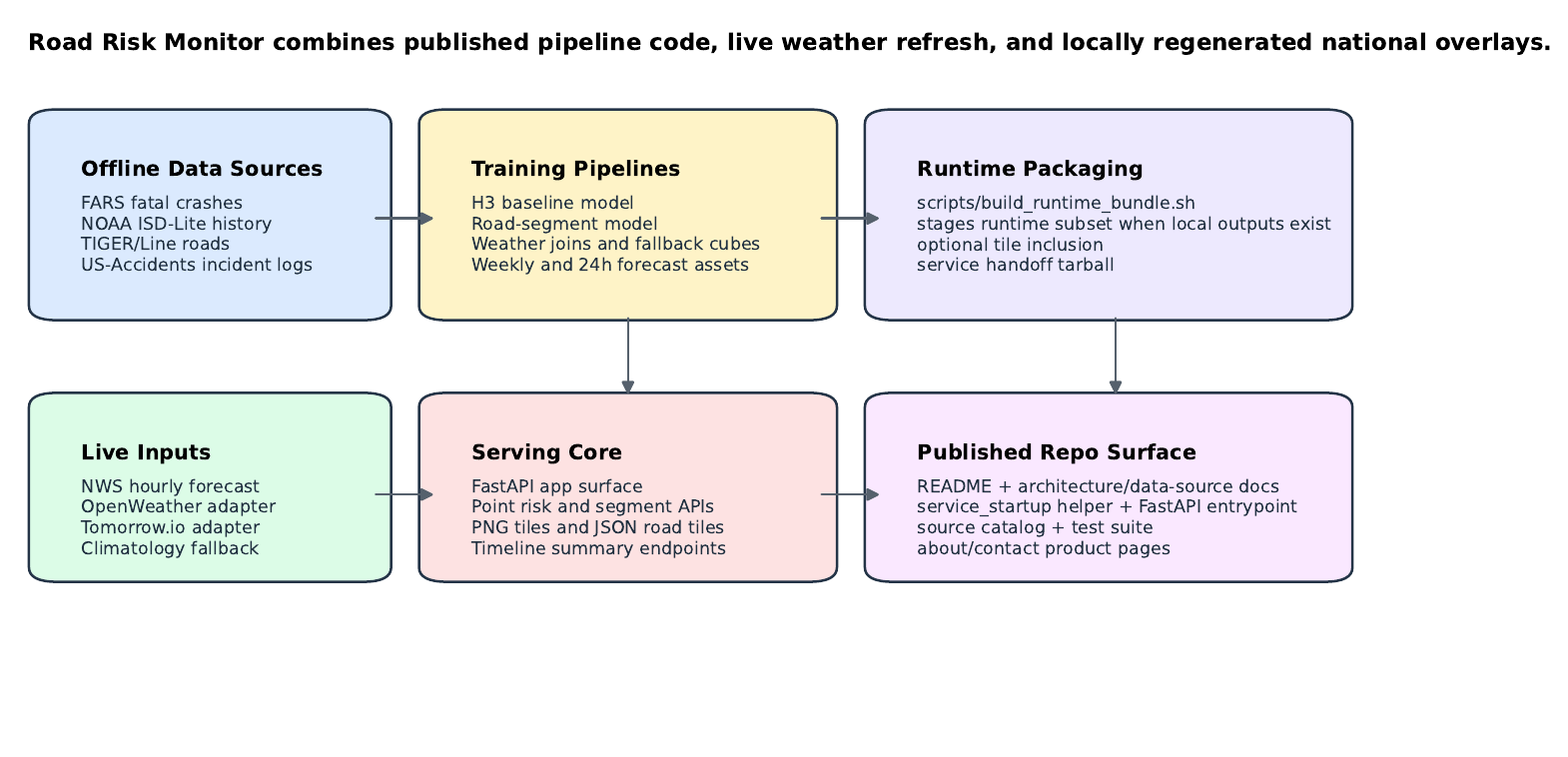}
  \caption{\productname{} is intentionally layered. The published repository contains the offline pipelines, serving endpoints, runtime-packaging utilities, and documentation needed to regenerate trained models and overlay artifacts locally.}
  \label{fig:overview}
\end{figure*}

Figure~\ref{fig:overview} summarizes the published repository. Offline data from FARS, NOAA ISD-Lite, TIGER/Line, and US-Accidents is transformed into training tables and serving assets. Online inputs use NWS first, with optional OpenWeather and Tomorrow.io adapters and climatology fallback when needed. The service then exposes predictions through a FastAPI application and provides a runtime-bundle script for handing a rebuilt artifact into an operational environment.

The system has two linked prediction layers. The first is a nationwide baseline overlay that scores H3 cells on a weekly cycle using historical fatal-crash intensity and representative-station weather climatology. The second is a road-level service that converts TIGER/Line geometry into short segments, matches incident history to those segments, scores them with weather-aware models, and refreshes a rolling \TrafficSafetyForecastFrames{}-hour forecast.

\section{Data Sources and Representations}
No single source is sufficient for this problem. The repository therefore uses several complementary public datasets and interfaces, summarized in Table~\ref{tab:sources}.

\begin{table*}[t]
  \centering
  \scriptsize
  \caption{Core sources used by \productname{}.}
  \label{tab:sources}
  \resizebox{\textwidth}{!}{%
  \begin{tabular}{p{1.45in} p{2.0in} p{1.85in} p{2.15in}}
    \toprule
    Source & Offline role & Online role & Why it matters \\
    \midrule
    FARS \citep{fars} & Baseline fatal-crash labels and historical H3 counts & None directly & Official national fatal-crash census with stable long-range coverage \\
    NOAA ISD-Lite \citep{noaaisdlite} & Historical hourly weather and representative-station climatology & Fallback weather context when live fetch fails & Reproducible weather features across the U.S. \\
    TIGER/Line roads \citep{tigerline} & Nationwide road geometry and segment construction & Supports road-level map tiles and detail views & Stable open national road network \\
    US-Accidents \citep{usaccidents} & Segment-event labels, severity, and event weather context & None directly in the live service & Denser road incident history than fatal-only data \\
    NWS API \citep{nwsapi} & Not used during offline fitting & Default live forecast source for point scoring and road refresh & Enables real +24-hour refresh without a paid weather contract \\
    H3 \citep{h3} & Baseline spatial index and candidate-cell representation & Stable lookup unit for overlay serving & Clean national aggregation unit for the baseline layer \\
    \bottomrule
  \end{tabular}
  }
\end{table*}

For the baseline model, the canonical training example is an H3 cell-hour feature vector. In our local rebuild from the published scripts, the baseline bundle uses H3 resolution \TrafficSafetyBaseResolution{}, \TrafficSafetyCandidateCells{} candidate cells, and weather climatology derived from \TrafficSafetyLiveStations{} representative stations. FARS incidents are cleaned, mapped to cells, converted to hour-of-week counts, and expanded through neighboring cells to create the candidate surface.

For the road model, the canonical example is a segment-hour feature vector. In the same rebuild, TIGER/Line primary and secondary roads are split into short segments, producing \TrafficSafetyRoadSegments{} total road segments and \TrafficSafetyActiveRoadSegments{} active segments in the live road layer. US-Accidents points are matched to those segments via BallTree candidate retrieval followed by exact point-to-polyline distance, yielding \TrafficSafetySegmentEvents{} matched events across \TrafficSafetyStatesCovered{} states. Matching quality is operationally plausible: median distance is \TrafficSafetySegmentMatchMedianM{} m, mean distance is \TrafficSafetySegmentMatchMeanM{} m, and the 95th percentile is \TrafficSafetySegmentMatchPctlNinetyFiveM{} m.

\section{Modeling and Serving Stack}
\subsection{Nationwide Baseline}
The nationwide baseline is trained from cleaned FARS events and representative-station weather. Its feature vector includes latitude, longitude, cyclical encodings of hour, day, and month, historical cell and same-hour counts, and weather covariates including temperature, dewpoint, humidity, wind speed, and a wet-hour indicator. The stored model uses \TrafficSafetyBaseFeatureCount{} features and reports \TrafficSafetyBaseAUC{} AUROC and \TrafficSafetyBaseAP{} average precision on its held-out year.

This baseline is not the whole product. Its role is to guarantee national coverage and a stable weekly overlay even when no road-level live signal is available.

\subsection{Road-Segment Forecasting}
The road-level model adds finer spatial resolution and live-weather adaptation. Each segment receives static geometry and road-class features together with temporal, historical, and weather covariates. The stored bundle uses \TrafficSafetySegmentFeatureCount{} features and reports \TrafficSafetySegmentAUC{} AUROC and \TrafficSafetySegmentAP{} average precision on an internal year holdout derived from the same matched-event pipeline.

That number needs careful interpretation. The baseline and segment evaluations are not comparable regimes. The baseline result is a held-out-year evaluation on the national H3 task, while the segment result comes from a same-pipeline train/eval setup with matched US-Accidents labels, sampled negatives, stable segment identities, and strong historical-count features. Under this construction, recurring high-activity segments can become very easy to separate from sampled non-events, so near-perfect discrimination can reflect task formulation and persistence effects rather than broad deployment validity.

For that reason, this paper does not use the segment metric as a headline claim. We treat it as a diagnostic signal that the current segment task is internally separable, while the right next evaluations are geographic holdouts, cold-segment holdouts, source holdouts, and prospective roll-forward tests against future live data. The important point for the present paper is that the segment model is embedded in a live service that can refresh a national 24-hour road forecast and expose that forecast through both raster and vector-style tiles.

\subsection{Application Surface}
The FastAPI app exposes both machine and human-facing routes. Excluding documentation routes, the repository currently serves \TrafficSafetyPublicRoutes{} public paths: \TrafficSafetyHtmlRoutes{} HTML product pages and \TrafficSafetyMachineRoutes{} machine-oriented endpoints. The machine surface includes health checks, live point risk scoring, road-segment queries, detail inspection, timeline summaries, and tile endpoints. The page surface includes the interactive map, an about page, and a contact flow with SMTP delivery when configured and local fallback logging otherwise.

\section{Runtime Packaging and Published Surface}
This repository goes beyond a research artifact by publishing an explicit runtime handoff path. The script \path{scripts/build_runtime_bundle.sh} stages code, scripts, locally generated models, optional tiles, and selected processed data into a runnable tarball when those artifacts exist on disk. The service entrypoint in \path{src/main.py} also imports \path{scripts/service_startup.py} so that startup behavior is codified rather than hidden in an external shell wrapper.

The published surface is broader than the runtime script alone. The repository includes \path{README.md}, \path{ARCHITECTURE.md}, \path{DATA_SOURCES.md}, and \path{source_catalog.json}. Together, those files document the rebuild path, source assumptions, and service shape. In other words, the public repo exposes the operational logic even though bulky local outputs such as downloaded data, trained models, and tiles are generated rather than versioned.

\section{Local Build Evidence}
\begin{figure*}[t]
  \centering
  \includegraphics[width=\textwidth]{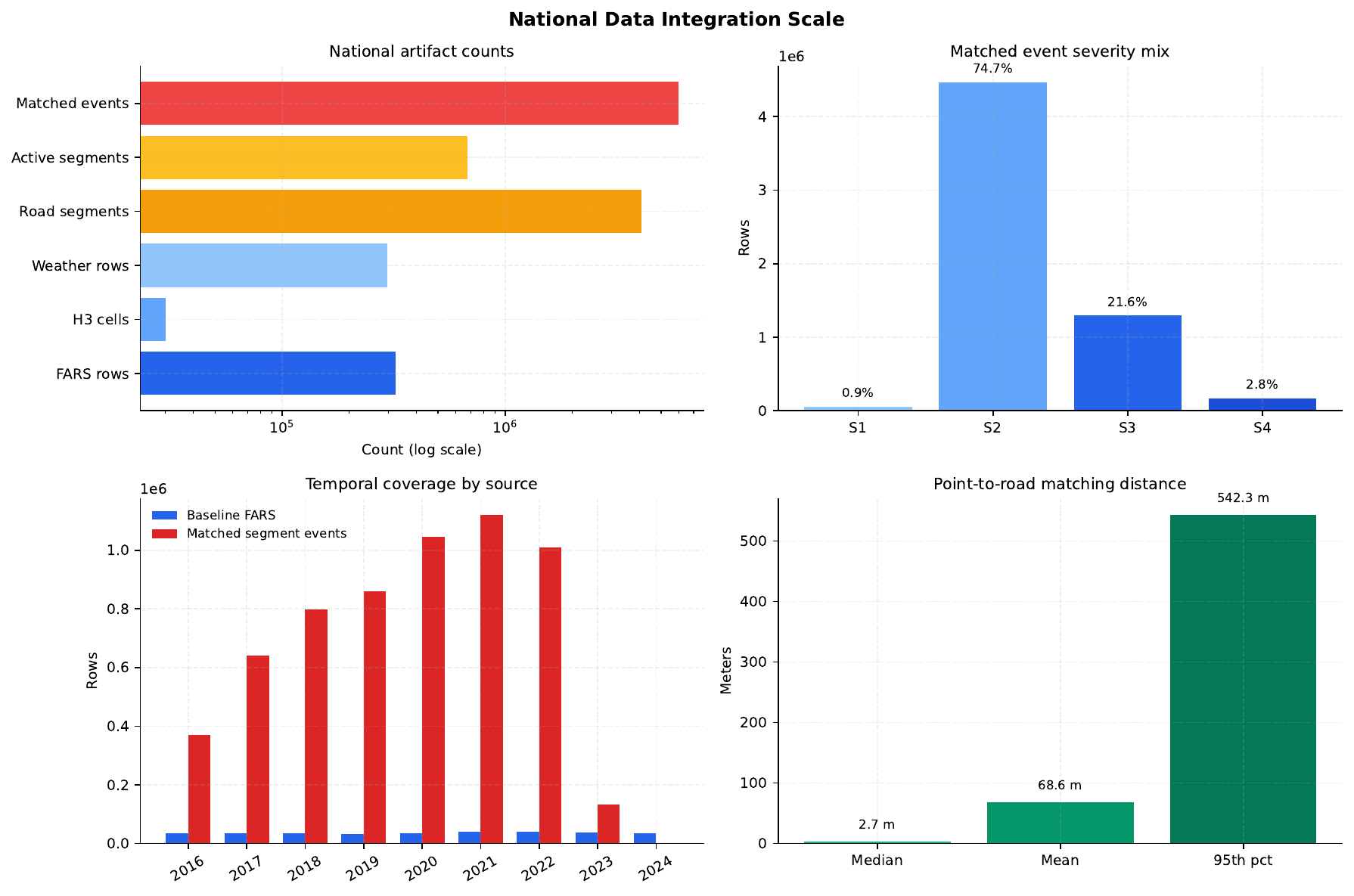}
  \caption{National data integration evidence from a local rebuild driven by the published repository: source counts, multi-year coverage, matched-event severity composition, and the geometry quality of point-to-road alignment.}
  \label{fig:scale}
\end{figure*}

\begin{figure*}[t]
  \centering
  \includegraphics[width=\textwidth]{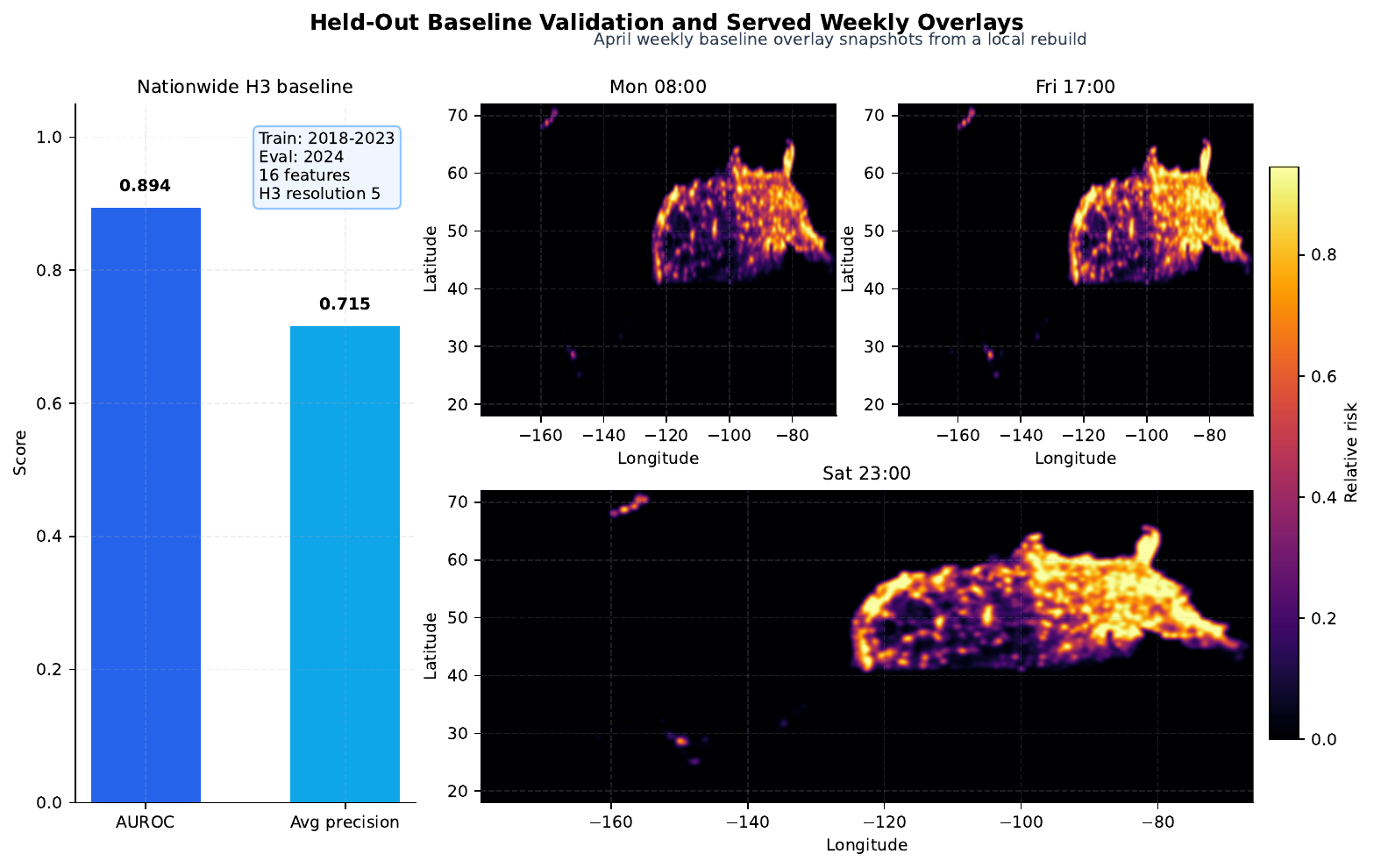}
\caption{Held-out baseline validation together with served nationwide overlay snapshots from a local rebuild. The maps show that the published code can be turned into actual spatiotemporal outputs used by the web product.}
  \label{fig:ops}
\end{figure*}

Table~\ref{tab:artifact} and Figures~\ref{fig:scale} and~\ref{fig:ops} quantify a local rebuild executed from the published repository in the shared \texttt{playground} environment rather than pre-versioned outputs in the Git tree.

First, the rebuild is national in scale. It produced \TrafficSafetyAccidentRows{} cleaned FARS incidents, \TrafficSafetyWeatherClimatologyRows{} weather rows, \TrafficSafetyRoadSegments{} road segments, and \TrafficSafetySegmentEvents{} matched segment events. Figure~\ref{fig:scale} makes that concrete through source counts, temporal coverage, severity composition, and point-to-road matching quality. Second, the rebuild does not stop at tables: it also produced served overlay artifacts. Figure~\ref{fig:ops} combines the held-out H3 baseline result with snapshots from the rebuilt weekly overlay tensor, which is the same kind of artifact the web application uses for nationwide playback. Finally, the evaluation story should still be read with care. The baseline metric is the cleaner scientific performance result, while the segment metric is retained only as an internal diagnostic because its split is materially easier.

\begin{table}[t]
  \centering
  \small
  \caption{Local rebuild data integration and validation summary.}
  \label{tab:artifact}
  \begin{tabular}{p{1.55in} p{1.45in}}
    \toprule
    Quantity & Value \\
    \midrule
    Cleaned FARS incidents & \TrafficSafetyAccidentRows{} \\
    Candidate H3 cells & \TrafficSafetyCandidateCells{} \\
    Weather climatology rows & \TrafficSafetyWeatherClimatologyRows{} \\
    Road segments & \TrafficSafetyRoadSegments{} \\
    Active road segments & \TrafficSafetyActiveRoadSegments{} \\
    Matched segment events & \TrafficSafetySegmentEvents{} \\
    States with matched events & \TrafficSafetyStatesCovered{} \\
    Match distance median / mean & \TrafficSafetySegmentMatchMedianM{} m / \TrafficSafetySegmentMatchMeanM{} m \\
    Match distance 95th pct & \TrafficSafetySegmentMatchPctlNinetyFiveM{} m \\
    Baseline AUROC / AP & \TrafficSafetyBaseAUC{} / \TrafficSafetyBaseAP{} \\
    Segment split & Internal same-pipeline year holdout \\
    Forecast horizon & \TrafficSafetyForecastFrames{} h \\
    Live NWS stations & \TrafficSafetyNWSStations{} / \TrafficSafetyLiveStations{} \\
    \bottomrule
  \end{tabular}
\end{table}

\section{Discussion}
The strongest aspect of \productname{} is integration. The H3 baseline alone would be too coarse for road operations, and the road-segment model alone would be difficult to operationalize without a national weather backbone, tile artifacts, and route surface. The repository is useful because it combines those pieces into one concrete system.

The deployability layer also changes the interpretation of the work. A repository that ships runtime packaging scripts, startup helpers, documentation, and tested public routes is materially closer to a production service than a repository that stops at notebooks and checkpoints. In that sense, the packaging and service code is part of the contribution, not peripheral engineering.

The segment metric deserves explicit caution. A reviewer seeing \TrafficSafetySegmentAUC{} AUROC on a road task is right to ask whether the split is too easy, whether leakage is possible through persistent segment identity and historical-count features, or whether sampled negatives make the task much cleaner than real deployment. Our view is that this is exactly why the number should not be marketed as the paper's scientific centerpiece. The correct next step is not to defend the number aggressively, but to run harder evaluations that better match real deployment.

At the same time, the limitations are clear. The baseline task is anchored on fatal crashes, the segment task relies on matched US-Accidents records and sampled negatives, and the live adaptation layer uses weather only rather than richer exposure, work-zone, or traffic-speed signals. Those are exactly the right next steps now that the train-to-serve pipeline already exists.

\section{Reproducibility}
Published code and documentation are available at \url{https://github.com/TonyIvchenko/traffic-safety}. The repository can be rebuilt from scripts rather than reconstructed from prose. The public documentation surface centers on \path{README.md}, \path{ARCHITECTURE.md}, \path{DATA_SOURCES.md}, and \path{source_catalog.json}. The offline stack then centers on:
\begin{itemize}[leftmargin=1.1em,itemsep=0.12em,topsep=0.2em]
\item \path{scripts/build_dataset.py}
\item \path{scripts/download_weather.py}
\item \path{scripts/train_model.py}
\item \path{scripts/build_segments.py}
\item \path{scripts/build_segment_events.py}
\item \path{scripts/train_segment_model.py}
\item \path{scripts/refresh_segment_tiles.py}
\end{itemize}

The serving surface is implemented in \path{src/main.py}, \path{src/live_weather.py}, and \path{src/road_tiles.py}; runtime handoff is documented in \path{scripts/build_runtime_bundle.sh} and \path{scripts/service_startup.py}. Large local outputs such as downloaded data, trained models, tiles, and runtime bundles are generated by the published scripts and are not part of the public Git tree. The published application surface is validated by \TrafficSafetySmokeTests{} smoke tests in \path{tests/test_main.py}; in this session they passed in the shared \texttt{playground} environment with \texttt{pytest -q -p no:cacheprovider tests/test\_main.py}.

\section{Conclusion}
This paper presented \productname{}, a multi-source national road-safety system that spans data ingestion, modeling, live weather adaptation, tile serving, runtime packaging, and service startup. The main result is not just that a local rebuild achieves useful validation metrics, but that the entire path from public road-safety data to a deployable public service is encoded in one published repository plus a reproducible rebuild path. For applied road-risk ML, that systems completeness is the contribution.

\bibliographystyle{plainnat}
\bibliography{references}

\begin{thebibliography}{6}
\providecommand{\natexlab}[1]{#1}
\providecommand{\url}[1]{\texttt{#1}}
\expandafter\ifx\csname urlstyle\endcsname\relax
  \providecommand{\doi}[1]{doi: #1}\else
  \providecommand{\doi}{doi: \begingroup \urlstyle{rm}\Url}\fi

\bibitem[{H3 Open Source Community}(2026)]{h3}
{H3 Open Source Community}.
\newblock {H3}: A hexagonal hierarchical geospatial indexing system.
\newblock \url{https://h3geo.org/}, 2026.
\newblock Accessed 2026-05-04.

\bibitem[Moosavi(2024)]{usaccidents}
M.~Moosavi.
\newblock {US-Accidents}: A countrywide traffic accident dataset.
\newblock \url{https://smoosavi.org/datasets/us_accidents}, 2024.
\newblock Accessed 2026-05-04.

\bibitem[{National Centers for Environmental Information}(2026)]{noaaisdlite}
{National Centers for Environmental Information}.
\newblock {NOAA} integrated surface database ({ISD}).
\newblock \url{https://www.ncei.noaa.gov/products/land-based-station/integrated-surface-database}, 2026.
\newblock Accessed 2026-05-04.

\bibitem[{National Highway Traffic Safety Administration}(2026)]{fars}
{National Highway Traffic Safety Administration}.
\newblock Fatality analysis reporting system ({FARS}).
\newblock \url{https://www.nhtsa.gov/research-data/fatality-analysis-reporting-system-fars}, 2026.
\newblock Accessed 2026-05-04.

\bibitem[{National Weather Service}(2026)]{nwsapi}
{National Weather Service}.
\newblock National weather service {API} web service documentation.
\newblock \url{https://www.weather.gov/documentation/services-web-api}, 2026.
\newblock Accessed 2026-05-04.

\bibitem[{United States Census Bureau}(2026)]{tigerline}
{United States Census Bureau}.
\newblock {TIGER/Line} shapefiles.
\newblock \url{https://www.census.gov/geographies/mapping-files/time-series/geo/tiger-line-file.html}, 2026.
\newblock Accessed 2026-05-04.

\end{thebibliography}

\end{document}